\pdfoutput=1

\documentclass[11pt]{article}

\usepackage[preprint]{acl}

\usepackage{times}
\usepackage{latexsym}

\usepackage[T1]{fontenc}

\usepackage[utf8]{inputenc}

\usepackage{microtype}

\usepackage{graphicx}
\usepackage{amsmath}

\usepackage{booktabs}
\usepackage{array}
\usepackage{tabularx}

\usepackage{multirow}

\title{\textit{Smarter, Better, Faster, Longer}: A Modern Bidirectional Encoder for Fast, Memory Efficient, and Long Context Finetuning and Inference}

\author{
 \textbf{Benjamin Warner\textsuperscript{1\textdagger}} \hspace{0.2em}
 \textbf{Antoine Chaffin\textsuperscript{2\textdagger}} \hspace{0.2em}
 \textbf{Benjamin Clavié\textsuperscript{1\textdagger}}
\\
 \textbf{Orion Weller\textsuperscript{3}} \hspace{0.2em}
 \textbf{Oskar Hallström\textsuperscript{2}} \hspace{0.2em}
 \textbf{Said Taghadouini\textsuperscript{2}}
\\
 \textbf{Alexis Gallagher\textsuperscript{1}} \hspace{0.2em}
 \textbf{Raja Biswas\textsuperscript{1}} \hspace{0.2em}
 \textbf{Faisal Ladhak\textsuperscript{4*}} \hspace{0.2em}
 \textbf{Tom Aarsen\textsuperscript{5}}
\\
 \textbf{Nathan Cooper\textsuperscript{1}} \hspace{0.2em}
 \textbf{Griffin Adams\textsuperscript{1}} \hspace{0.2em}
 \textbf{Jeremy Howard\textsuperscript{1}} \hspace{0.2em}
 \textbf{Iacopo Poli\textsuperscript{2}}
\\[1ex]
 \textsuperscript{1}Answer.AI
 \hspace{0.2em}
 \textsuperscript{2}LightOn
 \hspace{0.2em}
 \textsuperscript{3}Johns Hopkins University
 \hspace{0.2em}
 \textsuperscript{4}NVIDIA
 \hspace{0.2em}
\textsuperscript{5}HuggingFace
\\
 \small{
   \textit{\textdagger: core authors, *: work done while at Answer.AI}
   }
   \\
\small{
   \textbf{Correspondence:} \{bw,bc\}@answer.ai, antoine.chaffin@lighton.ai
 }
}

\newcommand\blfootnote[1]{%
  \begingroup
  \renewcommand\thefootnote{}\footnote{#1}%
  \addtocounter{footnote}{-1}%
  \endgroup
}

\begin{document}
\maketitle
\begin{abstract}
Encoder-only transformer models such as BERT offer a great performance-size tradeoff for retrieval and classification tasks with respect to larger decoder-only models. Despite being the workhorse of numerous production pipelines, there have been limited Pareto improvements to BERT since its release. In this paper, we introduce ModernBERT, bringing modern model optimizations to encoder-only models and representing a major Pareto improvement over older encoders. Trained on 2 trillion tokens with a native 8192 sequence length, ModernBERT models exhibit state-of-the-art results on a large pool of evaluations encompassing diverse classification tasks and both single and multi-vector retrieval on different domains (including code). In addition to strong downstream performance, ModernBERT is also the most speed and memory efficient encoder and is designed for inference on common GPUs. 
\end{abstract}

\section{Introduction}

After the release of BERT~\cite{bert}, encoder-only transformer-based~\cite{DBLP:conf/nips/VaswaniSPUJGKP17} language models dominated most applications of modern Natural Language Processing (NLP). Despite the rising popularity of Large Language Models (LLMs) such as GPT~\cite{Radford2018ImprovingLU,Radford2019,gpt3}, Llama~\cite{llama2,llama3}, and Qwen~\cite{qwen,qwen2}, encoder-only models remain widely used in a variety of non-generative downstream applications.\blfootnote{\href{https://github.com/AnswerDotAI/ModernBERT}{https://github.com/AnswerDotAI/ModernBERT}}

The encoder's popularity is largely due to their modest inference requirements, enabling them to efficiently process corpora of documents at scale for retrieval and quickly perform discriminative tasks. Encoder models offer a compelling trade-off in quality versus size, making them a popular option against encoder-decoder and decoder-only language models when dealing with substantial amounts of data~\cite{fineweb-edu}.

Encoder models are particularly popular in Information Retrieval (IR) applications, e.g., semantic search, with notable progress on leveraging encoders for this task~\cite{DBLP:conf/emnlp/KarpukhinOMLWEC20, DBLP:conf/sigir/KhattabZ20}. While LLMs have taken the spotlight in recent years, they have also motivated a renewed interest in encoder-only models for IR. Indeed, encoder-based semantic search is a core component of Retrieval-Augmented Generation (RAG) pipelines~\cite{rag}, where encoder models are used to retrieve and feed LLMs with context relevant to user queries.

Encoder-only models are also still frequently used for a variety of discriminative tasks such as classification~\cite{setfit} or Natural Entity Recognition (NER)~\cite{gliner}, where they often match the performance of specialized LLMs. Here again, they can be used in conjunction with LLMs, for example detecting toxic prompts~\cite{beavertails, wildjailbreak} and preventing responses, or routing queries in an agentic framework~\cite{react, toolformer}.

Surprisingly, these pipelines currently rely on older models, and quite often on the original BERT itself as their backbone~\cite{e5,bge}, without leveraging improvements developed in recent years. Practitioners face many drawbacks: sequence lengths limited to 512 tokens, suboptimal model design~\cite{codesigning} and vocabulary sizes~\cite{karpathy}, and generally inefficient architectures, whether in terms of downstream performance or computational efficiency. Finally, training data is limited in volume and restricted to narrow domains (especially lacking code data) or lacking knowledge of recent events. 

Recent modernization efforts have only partially addressed the shortcomings of encoder-only models due to limited breadth. MosaicBERT~\cite{mosaic}, CrammingBERT~\cite{crammingbert}, and AcademicBERT~\cite{academicbudget} focused on matching BERT performance with better training efficiency. NomicBERT~\cite{nomic} and GTE-en-MLM~\cite{gte} (developed concurrently to this work) introduced longer-context encoder models focused on retrieval applications, but did not optimize for efficiency or classification performance, and re-used older training data mixtures which is especially apparent in programming-related tasks. 

\textbf{Contributions} We present ModernBERT, a modernized encoder-only transformer model, with an improved architecture designed to increase downstream performance and efficiency, especially over longer sequence lengths. We also bring encoder-only models to modern, larger data scales, by training on 2 trillion tokens, with a data mixture including code data. We release two models, \href{https://huggingface.co/answerdotai/ModernBERT-base}{\textbf{ModernBERT-base}} and \href{https://huggingface.co/answerdotai/ModernBERT-large}{\textbf{ModernBERT-large}}, which reach state-of-the-art overall performance against all existing encoder models on a wide variety of downstream tasks. These results are achieved with considerably higher inference efficiency, processing sequences of 8192 tokens almost two times faster than previous models.

To support future research on encoder-only models, we release \href{https://github.com/AnswerDotAI/ModernBERT}{FlexBERT}\footnote{FlexBERT is built on top of a revised MosaicBERT~\cite{mosaic} codebase.}, our modular architecture framework allowing easy experimentation, and inspired by Pythia~\cite{pythia}, all intermediate training checkpoints (further detailed in Section~\ref{subsubsec:trainingsetting}).

\section{Methods}

\subsection{Architectural Improvements}
\label{sec:arch}

Our model architecture extends the standard transformer architecture~\cite{DBLP:conf/nips/VaswaniSPUJGKP17} by incorporating extensively tested recent advances (Section~\ref{sec:transformerplusplus}). We introduce additional efficiency-oriented modifications, through both architectural and implementation improvements (Section~\ref{sec:efficiency}) and a GPU optimized model design (Section~\ref{sec:modeldesign}). All of our architectural decisions were informed by ablations, which we detail in Appendix~\ref{sec:archablations}.

\subsubsection{Modern Transformer}
\label{sec:transformerplusplus}

\textbf{Bias Terms} Following~\cite{Dayma_DALLE_Mini_2021}, we disable bias terms in all linear layers except for the final decoder linear layer\footnote{While many efficient BERT training recipes disable the bias term in the decoder, e.g. ~\citet{crammingbert}, we hypothesized a decoder bias might help alleviate weight tying's negative effects~\cite{RepresentationDP,improvinglowcomputelanguage}.}. We also disable all bias terms in Layer Norms ~\cite{disablelayernormbias}. These two changes allow us to spend more of our parameter budget in linear layers.

\textbf{Positional Embeddings} We use rotary positional embeddings (RoPE)~\cite{DBLP:journals/ijon/SuALPBL24} instead of absolute positional embeddings. This choice is motivated by the proven performance of RoPE in short- and long-context language models~\cite{neox,llama3,gemma2}, efficient implementations in most frameworks, and ease of context extension. 

\textbf{Normalization} We use a pre-normalization block~\cite{DBLP:conf/icml/XiongYHZZXZLWL20} with the standard layer normalization~\cite{layernorm}, which is known to help stabilize training~\cite{DBLP:conf/icml/XiongYHZZXZLWL20}. Similar to CrammingBERT~\cite{crammingbert} which also uses pre-normalization, we add a LayerNorm after the embedding layer. To avoid repetition, we remove the first LayerNorm in the first attention layer.

\textbf{Activation} We adopt GeGLU~\cite{shazeerglu}, a Gated-Linear Units (GLU)-based~\cite{pmlr-v70-dauphin17a} activation function built on top of the original BERT's GeLU~\cite{hendrycks2016gaussian} activation function. This is in line with recent work showing consistent empirical improvements when using GLU variants~\cite{shazeerglu,crammingbert}.

\subsubsection{Efficiency Improvements}
\label{sec:efficiency}

\textbf{Alternating Attention} Following recent work on efficient long context models~\cite{gemma2}, attention layers in ModernBERT alternate between global attention, where every token within a sequence attends to every other token, and local attention, where tokens only attend to each other within a small sliding window~\cite{longformer}. In ModernBERT, every third layer employs global attention with a RoPE theta of 160,000 and the remaining layers use a 128 token, local sliding window attention with a RoPE theta of 10,000.

\textbf{Unpadding} ModernBERT follows MosaicBERT~\cite{mosaic} and GTE~\cite{gte} in employing unpadding~\cite{unpadding} for both training and inference. Encoder-only language models typically use padding tokens to ensure a uniform sequence length in a batch, wasting compute on semantically empty tokens. Unpadding avoids this inefficiency by removing padding tokens, concatenating all sequences from a minibatch into a single sequence, and processing it as a batch of one. Prior unpadding implementations unpad and repad sequences internally for different model layers, wasting compute and memory bandwidth. We use Flash Attention's variable length attention and RoPE implementations, allowing jagged attention masks and RoPE applications on one unpadded sequence. ModernBERT unpads inputs before the token embedding layer and optionally repads model outputs leading to a 10-to-20 percent performance improvement over other unpadding methods.

\textbf{Flash Attention} Flash Attention~\cite{FA} is a core component of modern transformer-based models, providing memory and compute efficient attention kernels. At the start of this work, Flash Attention 3~\cite{flash-attention-3}, the most recent iteration for Nvidia H100 GPUs, did not include support for sliding window attention. ModernBERT uses a mixture of Flash Attention 3 for global attention layers and Flash Attention 2~\cite{FA2} for local attention layers.  

\textbf{torch.compile} We leverage PyTorch's built-in compiling~\cite{ansel2024pytorch} to improve the training efficiency by compiling all compatible modules. This yields a 10 percent improvement in throughput with negligible compilation overhead.

\subsubsection{Model Design}
\label{sec:modeldesign}

At the same parameter count, models with more narrow layers (\textit{Deep \& Narrow}) have different learning patterns than models with fewer wide layers (\textit{Shallow \& Wide})~\cite{nguyen2021do}.~\citet{efficientT5} and~\cite{mobilellm} have shown that \textit{Deep \& Narrow} language models have better downstream performance than their shallower counterparts, at the expense of slower inference.

~\citet{codesigning} highlighted that large runtime gains can be unlocked by designing models in a \textit{hardware-aware} way, which had previously been anecdotally observed by many practitioners~\cite{megatron,karpathy,neox}. ModernBERT was designed through many small-scale ablations to maximize the utilization of a basket of common GPUs\footnote{Which, at the time of this work, are server GPUs: NVIDIA T4, A10, L4, A100, and H100 and consumer GPUs: NVIDIA RTX 3090 and 4090. Prioritization was given to inference GPUs (excluding A100 \& H100).}, while aiming to be as \textit{Deep \& Narrow} as possible without a significant inference slowdown.

ModernBERT has 22 and 28 layers for the base and large models, for a total parameter count of 149 and 395 million, respectively, striking the balance between downstream performance and hardware efficiency. ModernBERT base has a hidden size of 768 with a GLU expansion of 2,304, while large has a hidden size of 1,024 and GLU expansion of 5,248. These ratios allow optimal tiling across tensor cores and the most efficient tiling across the differing number of streaming multiprocessors on our target basket of GPUs. More details on model design are provided in Appendix~\ref{app:modeldesign}.

\subsection{Training}

\subsubsection{Data}
\textbf{Mixture} Both ModernBERT models are trained on 2 trillion tokens of primarily English data from a variety of data sources, including web documents, code, and scientific literature, following common modern data mixtures. We choose the final data mixture based on a series of ablations.

\textbf{Tokenizer} Unlike the majority of recent encoders which reuse the original BERT tokenizer~\cite{nomic,mosaic,gte}, we opt to use a modern BPE tokenizer. We use a modified version of the OLMo tokenizer ~\cite{olmo} which provides better token efficiency and performance on code-related tasks. The ModernBERT tokenizer uses the same special tokens (e.g., \textsc{[CLS]} and \textsc{[SEP]}) and templating as the original BERT model~\cite{bert}, facilitating backwards compatibility. To ensure optimal GPU utilization~\cite{codesigning,karpathy}, the vocabulary is set to 50,368, a multiple of 64 and includes 83 unused tokens to support downstream applications.

\textbf{Sequence Packing} In order to avoid high minibatch-size variance within our training batches as a result of unpadding, we adopt sequence packing~\cite{t5,seqpack} with a greedy algorithm, which resulted in a sequence packing efficiency of over 99 percent, ensuring batch size uniformity.

\subsubsection{Training Settings}
\label{subsubsec:trainingsetting}
\textbf{MLM} We follow the Masked Language Modeling (MLM) setup used by MosaicBERT~\cite{mosaic}. We remove the Next-Sentence Prediction objective which introduces noticeable overhead for no performance improvement~\cite{roberta,academicbudget}, and use a masking rate of 30 percent, as the original rate of 15 percent has since been shown to be sub-optimal~\cite{mask15maskedlanguage}.

\textbf{Optimizer} We use the StableAdamW optimizer~\cite{stableladamw}, which improves upon AdamW~\cite{adamw} by adding Adafactor-style~\cite{adafactor} update clipping as a per-parameter learning rate adjustment. StableAdamW's learning rate clipping outperformed standard gradient clipping on downstream tasks and led to more stable training. Hyperparameters details are given in Appendix~\ref{app:trainingsettings}.

\textbf{Learning Rate Schedule} During pretraining, we use a modified trapezoidal Learning Rate (LR) schedule~\cite{xing2018walksgd}, also known as Warmup-Stable-Decay (WSD)~\cite{DBLP:conf/cvpr/Zhai0HB22,DBLP:journals/corr/abs-2404-06395}. After a short LR warmup, the trapezoidal schedule holds the LR constant for the majority of training, followed by a short LR decay. This schedule has been shown to match the performance of cosine scheduling~\cite{DBLP:journals/corr/abs-2405-18392,mambaoutai} with the benefit of enabling continual training on any checkpoint without cold restart issues~\cite{DBLP:journals/corr/abs-1910-08475}. Unlike most trapezoidal schedules, we use a $1-\text{sqrt}$ LR decay~\cite{1minussqrt}, as we found it to outperform linear and cosine decay.

We trained ModernBERT-base at a constant LR of 8e-4 for 1.7 trillion tokens following a 3 billion token warmup. After a 2 billion token warmup, we trained ModernBERT-large at a LR of 5e-4 for 900 billion tokens. We rolled back and restarted training at 5e-5 for the remaining 800 billion tokens after large’s loss plateaued for a few hundred billion tokens at 5e-4.

\textbf{Batch Size Schedule} Batch size scheduling starts with smaller gradient accumulated batches, increasing over time to the full batch size. In ablations, this schedule accelerated training progress. We warmup the batch size from 768 to 4,608 over 50 billion tokens and from 448 to 4,928 over 10 billion tokens, for ModernBERT-base and -large, respectively, with an uneven token schedule so each batch size has the same number of update steps. Details are provided in Appendix~\ref{subsec:batchsizewarmup}.

\textbf{Weight Initialization and Tiling}
We initialize ModernBERT-base with random weights following the Megatron initialization~\cite{megatron}. For ModernBERT-large, we follow the Phi model family~\cite{phi15, phi2}\footnote{As detailed in their 2023 NeurIPS presentation.} and initialize -large’s weights from ModernBERT-base. In ablation runs, this consistently matched Phi's improved training results and greatly speed up the initial loss decrease of our model training\footnote{This initialization reduced the amount of batch size and LR warmup needed for ModernBERT-large}. Details are provided in Appendix~\ref{subsec:weightiling}.

\textbf{Context Length Extension}
After training on 1.7 trillion tokens at a 1024 sequence length and RoPE theta of 10,000, we extend the native context length of ModernBERT to 8192 tokens by increasing the global attention layer’s RoPE theta to 160,000 and train for an additional 300 billion tokens. We first train at a constant lower learning rate\footnote{We only lowered the LR for ModernBERT-base, as large already decreased LR during the 1024 token training phase.} of 3e-4 for 250 billion tokens on an 8192 token mixture of the original pretraining dataset sampled following ~\citet{dataengineering}. Next, we upsample higher-quality sources following~\citet{prolong} and conduct the decay phase with a $1-\text{sqrt}$ LR schedule over 50 billion tokens. This context extension process yielded the most balanced model on downstream tasks, as most of our ablations using only one of these strategies resulted in a performance loss on either retrieval or classification tasks.

\section{Downstream Evaluation}
We performed an extensive set of evaluations, across a large range of tasks, aiming to demonstrate the versatility of ModernBERT in common scenarios.

For all tasks, ModernBERT is evaluated against existing encoders of similar size. The \textsc{base} size, conventionally defined as under 150 million parameters, includes \href{https://huggingface.co/google-bert/bert-base-uncased}{BERT-base}~\cite{bert}, \href{https://huggingface.co/microsoft/deberta-v3-base}{DeBERTa-v3-base}~\cite{debertav3}, \href{https://huggingface.co/FacebookAI/roberta-base}{RoBERTa-base}~\cite{roberta}, as well as the more recent 8192 context \href{https://huggingface.co/nomic-ai/NomicBERT-2048}{NomicBERT}~\cite{nomic} and \href{https://huggingface.co/Alibaba-NLP/GTE-en-MLM-base}{GTE-en-MLM-base}~\cite{gte}. The \textsc{large} size, conventionally defined as above 300 million and under 500 million parameters, includes \href{https://huggingface.co/google-bert/bert-large-uncased}{BERT-large-uncased}~\cite{bert}, \href{https://huggingface.co/microsoft/deberta-v3-large}{DeBERTa-v3-large}~\cite{debertav3} and \href{https://huggingface.co/FacebookAI/roberta-large}{RoBERTa-large}~\cite{roberta} and \href{https://huggingface.co/Alibaba-NLP/GTE-en-MLM-large}{GTE-en-MLM-large}~\cite{gte}. 

\subsection{Evaluation Setting}

\subsubsection{Natural Language Understanding}
\label{subsubsec:evalglue}

The General Language Understanding Evaluation (GLUE) benchmark~\cite{wang-etal-2018-glue} is the standard Natural Language Understanding (NLU) benchmark for encoder models, aiming to measure how well a model performs across a range of sentence or sentence-pair understanding tasks, such as sentiment detection~\cite{sst2} or language entailment, through tasks such as MNLI~\cite{MNLI}. Although GLUE is often regarded as saturated by the best-performing models, such as large language models~\cite{llmsurvey}, it remains one of the most commonly used evaluation suites for smaller encoder-based models, and provides a good impression of a model's performance on common classification tasks~\cite{mosaic,gte,debertav3}. 

We follow the practice of previous studies~\cite{bert,roberta,debertav3} and conduct a hyperparameter search on each GLUE subset (detailed in Appendix~\ref{subsec:fullglueresults}) in order to provide values comparable to other models.\footnote{As~\cite{gte} do not explicitly mention a parameter sweep, we initially ran the same hyperparameter sweep as we did for ModernBERT, but observed inconsistencies in the results. To avoid under-representing GTE-en-MLM's capabilities, we choose to use their reported GLUE results.}

\subsubsection{Text Retrieval}
\label{subsubsec:dpr}

Information Retrieval (IR) is one of the most common applications of encoder-only models,\footnote{At the time of this paper's writing, over half of the 100 most downloaded models on the \href{https://huggingface.co/models}{HuggingFace Model Hub} were encoder-based retrieval models.} where they are used to represent documents and queries in semantic search~\cite{DBLP:conf/emnlp/KarpukhinOMLWEC20}. This domain has recently seen considerable growth and interest following the spread of LLMs where semantic search powered by lightweight models is used to provide relevant context to LLMs as part of Retrieval-Augmented Generation pipelines.

We evaluate models in both the single-vector Dense Passage Retrieval (DPR)~\cite{DBLP:conf/emnlp/KarpukhinOMLWEC20} setting and the multi-vector ColBERT~\cite{DBLP:conf/sigir/KhattabZ20} setting. 

We report retrieval results on the popular BEIR evaluation suite~\cite{BEIR}, the common standard for evaluating retrieval performance across a variety of tasks and domains, using the nDCG@10 metric. For each setting detailed below, we conduct a learning rate sweep based on results over a subset of the BEIR benchmarks to select the final model, detailed in Appendix~\ref{subsec:fullBEIRresults}.

\textbf{Single vector retrieval}
One of the most common approaches to neural retrieval using encoders is DPR~\cite{DBLP:conf/emnlp/KarpukhinOMLWEC20}, where a single-vector is used to represent an entire document. The similarity between a query and a document can then be computed through distance operations, such as cosine similarity. Models are finetuned using contrastive learning to create representations which are close if a document is relevant to a query, and distant if not~\cite{DBLP:journals/corr/abs-1807-03748}. 

We train every base model using the MS-MARCO~\cite{msmarco} dataset with \href{https://huggingface.co/datasets/sentence-transformers/msmarco-co-condenser-margin-mse-sym-mnrl-mean-v1}{mined hard negatives}~\cite{hardnegs} on 1.25M samples with a batch size of 16 and learning rate warmup for 5\% of the training using \href{https://sbert.net/}{sentence-transformers}~\cite{reimers-2019-sentence-bert}.

\textbf{Multi vector retrieval}
\label{subsubsection:colbert}
Multi-vector retrieval, championed by ColBERT~\cite{DBLP:conf/sigir/KhattabZ20}, seeks to mitigate lost information from compressing an entire sequence into a single vector. In multi-vector retrieval, each document is represented by all of its individual token vectors, and the similarity between a query and a document is computed using the MaxSim\footnote{The sum for every query token of its similarity with the most similar document token} operator.

We adopt the training setup of JaColBERTv2.5~\cite{jacolbertv25}, an update on the ColBERTv2~\cite{DBLP:conf/naacl/SanthanamKSPZ22} training procedure, with a batch size of 16 and a 5\% learning rate warmup. We train all models by distilling the knowledge of a teacher model by using the KL-Divergence between the normalized teacher and student scores. Models are trained on \href{https://huggingface.co/datasets/lightonai/ms-marco-en-bge}{810k samples from MS-Marco}~\cite{msmarco} and teacher scores from BGE-M3~\cite{DBLP:conf/acl/ChenXZLLL24}, using the \href{https://github.com/lightonai/pylate}{PyLate} library~\cite{PyLate}.

\begin{table*}[!htbp]
\centering
\resizebox{\textwidth}{!}{
\begin{tabular}{llcccccccc}
 \toprule
 & & \multicolumn{3}{c}{IR (DPR)} & \multicolumn{2}{c}{IR (ColBERT)} & \multicolumn{1}{c}{NLU} & \multicolumn{2}{c}{Code} \\
 \cmidrule(l){3-5} \cmidrule(l){6-7} \cmidrule(l){8-8} \cmidrule(l){9-10}
& Model & \multicolumn{1}{c}{BEIR} & \begin{tabular}[c]{@{}c@{}}MLDR\textsubscript{OOD}\end{tabular} & \begin{tabular}[c]{@{}c@{}}MLDR\textsubscript{ID}\end{tabular} & BEIR & MLDR\textsubscript{OOD} & \begin{tabular}[c]{@{}c@{}}GLUE\end{tabular} & \begin{tabular}[c]{@{}c@{}}CSN\end{tabular} & \begin{tabular}[c]{@{}c@{}}SQA\end{tabular} \\

\midrule
\parbox[t]{3mm}{\multirow{6}{*}{\rotatebox[origin=c]{90}{
\parbox[c]{3cm}{\centering Base}}}}  
& BERT & 38.9 & 23.9 & 32.2 & 49.0 & 28.1 & 84.7 & 41.2 & 59.5  \\
& RoBERTa & 37.7 & 22.9 & 32.8 & 48.7 & 28.2 & 86.4 & 44.3 & 59.6  \\
& DeBERTaV3 & 20.2 & 5.4 & 13.4 & 47.1 & 21.9 & 88.1 & 17.5 & 18.6  \\
& NomicBERT & 41.0 & 26.7 & 30.3 & 49.9 & 61.3 & 84.0 & 41.6 & 61.4  \\
& GTE-en-MLM & 41.4 & \textbf{34.3} & \textbf{44.4} & 48.2 & 69.3 & 85.6 & 44.9 & 71.4  \\ 
& ModernBERT & \textbf{41.6} & 27.4 & 44.0 & \textbf{51.3} & \textbf{80.2} & \textbf{88.4} & \textbf{56.4} & \textbf{73.6}  \\
\midrule
\parbox[t]{3mm}{\multirow{5}{*}{\rotatebox[origin=c]{90}{
\parbox[c]{2cm}{\centering Large}}}} 
& BERT & 38.9 & 23.3 & 31.7 & 49.5 & 28.5 & 85.2 & 41.6 & 60.8  \\
& RoBERTa & 41.4 & 22.6 & 36.1 & 49.8 & 28.8 & 88.9 & 47.3 & 68.1  \\
& DeBERTaV3 & 25.6 & 7.1 & 19.2 & 46.7 & 23.0 & \textbf{91.4} & 21.2 & 19.7  \\
& GTE-en-MLM & 42.5 & \textbf{36.4} & \textbf{48.9} & 50.7 & 71.3 & 87.6 & 40.5 & 66.9 \\
& ModernBERT & \textbf{44.0} & 34.3 & 48.6 & \textbf{52.4} & \textbf{80.4} & 90.4 & \textbf{59.5} & \textbf{83.9}  \\
\bottomrule
\end{tabular}
}
\caption{Results for all models across an overview of all tasks. CSN refers to CodeSearchNet and SQA to StackQA. MLDR\textsubscript{ID} refers to in-domain (fine-tuned on the training set) evaluation, and MLDR\textsubscript{OOD} to out-of-domain.}
\label{tab:mainresults}
\end{table*}

\subsubsection{Long-Context Text Retrieval}

With a native 8192 context length, ModernBERT improves long-context performance over most existing encoders. However, there are relatively few standardized long-context benchmarks for encoder-only models, and most benchmarks, such as Needle-in-a-haystack~\cite{niah} and RULER~\cite{ruler} are geared towards generative tasks. Given this limitation, we demonstrate improved long-context performance on the English subset of \href{https://huggingface.co/datasets/Shitao/MLDR}{MLDR}~\cite{DBLP:conf/acl/ChenXZLLL24}, a long-context retrieval benchmark comprised of over 200,000 long documents. We evaluate three settings:

\textbf{Single Vector -- Out-Of-Domain } Models are trained on short-context MS-MARCO as described above, and is evaluated on long context MLDR without any further fine-tuning.

\textbf{Single Vector -- In Domain} Models trained on MS-MARCO are further fine-tuned on long-context MLDR training set before being evaluated.

\textbf{Multi-Vector -- Out-Of-Domain} Due to its token-level MaxSim mechanism, ColBERT models are able to generalize to long-context without any specific training~\cite{vespalongcolbert}. We directly evaluate the best checkpoints from Section~\ref{subsubsection:colbert} without any further fine-tuning on MLDR.

\subsubsection{Code Retrieval}

Fueled by increasingly good code completion models~\cite{llmcodesurvey}, downstream applications have quickly grown in popularity following the emergence of code assistants.\footnote{Spearheaded by \href{https://github.com/features/copilot}{GitHub Copilot} in 2021} Encoder-only models are used to process and retrieve large quantities of code-related information under resource constraints, increasing the importance of measuring and improving code capabilities of encoder models ~\cite{coir}.  Unlike most previous encoders which were largely trained only on textual data~\cite{bert,roberta,mosaic,gte,nomic}, ModernBERT is pre-trained on code and uses a code-aware tokenizer\footnote{Avoiding issues such as the ones seen in T5~\cite{t5}, whose vocabulary did not include curly braces.}. 

To measure programming-related performance, we evaluate all models on CodeSearchNet~\cite{codesearchnet}, a code-to-text benchmark where the model must identify relevant docstring or comments for code blocks, and StackOverflow-QA~\cite{coir}, where the model must identify relevant responses to StackOverflow questions, in a "hybrid" setting where documents contain both text and code. The latter benchmark also leverages long-context capabilities, as its queries and documents respectively contain 1,400 and 1,200 words on average, leading to average token counts of over 2000.

We evaluate these benchmarks using the CoIR (CodeIR) framework~\cite{coir}, as single-vector retrieval tasks. All models are trained by re-using the best hyper-parameters identified in Section~\ref{subsubsec:dpr}.

\begin{table*}[!t]
\centering
\begin{tabular}{llrrrrrrr}
\toprule
& & & \multicolumn{3}{c}{Short} & \multicolumn{3}{c}{Long} \\
\cmidrule(lr){4-6} \cmidrule(lr){7-9}
& Model & Params & BS & Fixed & Variable & BS & Fixed & Variable \\
\midrule
\parbox[t]{3mm}{\multirow{6}{*}{\rotatebox[origin=c]{90}{Base}}}
& BERT & 110M & 1096 & \textbf{180.4} & 90.2 & -- & -- & -- \\
& RoBERTa & 125M & 664 & 179.9 & 89.9 & -- & -- & -- \\
& DeBERTaV3 & 183M & 236 & 70.2 & 35.1 & -- & -- & -- \\
& NomicBERT & 137M & 588 & 117.1 & 58.5 & 36 & 46.1 & 23.1 \\
& GTE-en-MLM & 137M & 640 & 123.7 & 61.8 & 38 & 46.8 & 23.4 \\
& GTE-en-MLM\textsubscript{xformers} & 137M & 640 & 122.5 & 128.6 & 38 & 47.5 & 67.3 \\
& ModernBERT & 149M & \textbf{1604} & 148.1 & \textbf{147.3} & \textbf{98} & \textbf{123.7} & \textbf{133.8} \\
\midrule
\parbox[t]{3mm}{\multirow{5}{*}{\rotatebox[origin=c]{90}{Large}}}
& BERT & 330M & \textbf{792} & \textbf{54.4} & 27.2 & -- & -- & -- \\
& RoBERTa & 355M & 460 & 42.0 & 21.0 & -- & -- & -- \\
& DeBERTaV3 & 434M & 134 & 24.6 & 12.3 & -- & -- & -- \\
& GTE-en-MLM & 435M & 472 & 38.7 & 19.3 & 28 & 16.2 & 8.1 \\
& GTE-en-MLM\textsubscript{xformers} & 435M & 472 & 38.5 & 40.4 & 28 & 16.5 & 22.8 \\
& ModernBERT & 395M & 770 & 52.3 & \textbf{52.9} & \textbf{48} & \textbf{46.8} & \textbf{49.8} \\
\bottomrule
\end{tabular}
\caption{Memory (max batch size, \textit{BS}) and Inference (in thousands of tokens per second) efficiency results  on an NVIDIA RTX 4090, averaged over 10 runs. Dashes indicate unsupported configurations.}
\label{tab:efficiency}
\end{table*}

\subsection{Downstream Results and Discussion}

Aggregated results for all evaluations are presented in Table~\ref{tab:mainresults}. For BEIR and GLUE, the two common evaluation suites, we follow existing practice in reporting the average results. Detailed results are provided in Appendix~\ref{app:fullresults}.

In terms of downstream performance, ModernBERT is the strongest overall model at both the \textsc{base} and \textsc{large} model sizes. ModernBERT represents a Pareto improvement on all tasks over the original BERT and RoBERTA models, with better performance on every evaluation category.

\textbf{Short-Context Retrieval} On BEIR, both variants of ModernBERT outperform existing encoders in both the DPR and ColBERT settings, including the recent GTE-en-MLM and NomicBERT models designed to serve as better backbones for retrieval~\cite{gte,nomic}. 

While ModernBERT-base only narrowly edges out GTE-en-MLM-base on DPR evaluations, ModernBERT-large increases its lead despite having comparatively fewer parameters at 395M to GTE-en-MLM-large's 435M.

\textbf{Long-Context Retrieval - Single Vector} In the DPR setting, ModernBERT achieves impressive performance on MLDR, a long-context text retrieval task. However, these results also highlight an interesting phenomenon: without long-context finetuning ModernBERT outperforms both shorter-context models and the long-context NomicBERT but performs noticeably worse than GTE-en-MLM. The performance gap narrows considerably when evaluated in-domain, with both models performing similarly. This suggests that ModernBERT can effectively process long context sequences as a dense encoder but may require more adapted tuning. We plan to explore multiple potential explanations for this phenomenon in future work, including the impact of local attention or GTE-en-MLM having spent a larger part of its pretraining compute budget on longer sequence lengths~\cite{gte}.

\textbf{Long-Context Retrieval - Multi-Vector} In the ColBERT setting, long-context models (GTE-en-MLM, NomicBERT, and ModernBERT) all outperform short-context models by at least 40 NDCG@10 points without requiring any specific finetuning. These results confirm the findings of ~\citet{vespalongcolbert}, who showed that ColBERT models are particularly well-suited to long-context retrieval tasks. Among the long-context models, ModernBERT outperforms other long-context models, with at least a 9 NDCG@10 point lead on both model sizes. We theorize that these sizable gains could be explained by our long pretraining ensuring few, if any, tokens are under-trained, as well as a potentially synergistic effect of local attention with ColBERT-style retrieval, but leave further exploration of this phenomenon to future work.

\textbf{Natural Language Understanding} Both ModernBERT models demonstrate exceptional NLU results, as measured by GLUE. ModernBERT-base surpasses all existing base models, including DeBERTaV3-base, becoming the first MLM-trained model to do so. This is surprising, as DeBERTaV3 was trained with the Replaced-Token-Detection objective, which was previously thought to yield stronger downstream NLU performance~\cite{electra,debertav3}. ModernBERT-large is the second-best large encoder on GLUE, almost matching DeBERTaV3-large with one-tenth fewer parameters while processing tokens in half the time (see Section~\ref{sec:efficiencyresults}). 

\textbf{Code} On programming tasks, in both code-to-text (CodeSearchNet) and longer-context hybrid settings (StackQA), ModernBERT outperforms all other models. This result was expected, as it is the only evaluated encoder to be trained on a data mixture including programming data. These results, combined with ModernBERT's strong showings on other tasks, indicates that ModernBERT has improved understanding of code at no detriment to its ability to process natural text.

\section{Efficiency}
\label{sec:efficiencyresults}

\subsection{Evaluation Setting}

To measure inference efficiency across multiple sequence lengths, we create 4 synthetic sets of 8192 documents\footnote{Many common benchmarks are biased towards low and uniform sequence lengths, which is unrepresentative of many real-world situations.}.
The first two document sets are fixed-length: in \textit{fixed short-context}, all documents contain 512 tokens and in \textit{fixed long-context} all documents contain 8192 tokens\footnote{512 being the maximum length of most existing encoders, while 8192 is the maximum length of all long-context ones.}. To account for the impact of unpadding, we also create two varying-length document sets, where the number of tokens in each set are defined by a normal distribution centered on half the maximum sequence length, 256 and 4096 tokens, respectively. Full data statistics are provided in Appendix~\ref{app:efficiency}.

We then evaluate all models based on the number of tokens they can process per second, averaged over ten runs. All efficiency evaluations are ran on a single NVIDIA RTX 4090, one of the target GPUs of ModernBERT outlined in Section~\ref{sec:modeldesign}
We evaluate the GTE-en-MLM models under two settings: out-of-the box, and with the use of the xformers~\cite{xformers} library, which enables efficiency enhancements such as unpadding.
\subsection{Results}

All tokens-per-second efficiency results are presented in Table~\ref{tab:efficiency}, with absolute run-times provided in Appendix~\ref{app:efficiency}. ModernBERT stands out as the most efficient model overall. On short context, it processes fixed-length 512 token inputs faster than all other recent encoders, although slower than the original BERT and RoBERTa models\footnote{This is partially due to the relatively low parameter count of BERT and RoBERTa compared to more recent encoders.}. On long-context, ModernBERT is faster than all competing encoders, processing documents 2.65 and 3 times faster than the next-fastest encoder at the \textsc{base} and \textsc{large} sizes, respectively. ModernBERT-large's processing speed at length 8192 (46,801 tokens per second) is closer to that of GTE-en-MLM base (47,507 tokens per second) than it is to GTE-en-MLM-large (16,532 tokens per second).

On variable-length inputs, both GTE-en-MLM and ModernBERT models are considerably faster than all other models, largely due to unpadding. However, ModernBERT remains noticeably more efficient than GTE-en-MLM, processing 14.5-30.9 percent more tokens per second at low context lengths and 98.8-118.8 percent more at longer context lengths, thanks to its use of local attention.

ModernBERT is the overall most memory efficient model on both model sizes. ModernBERT-base is able to process batch sizes twice as large as every other model on both input lengths. ModernBERT-large is slightly less memory efficient than the original BERT-large on short-context inputs, but can process batches at least 60 percent bigger than every other large model.

\section{Conclusion}

We present ModernBERT, an open family of encoder-only models which set a new state of the art over existing encoder models on a wide range of classification and retrieval tasks. We show that encoders benefit from both recent pretraining data scales and architecture improvements from autoregressive LLMs.

ModernBERT has a native sequence length of 8,192 tokens and incorporates recent architecture improvements, such as GeGLU layers, RoPE positional embeddings, and alternating local-global attention. ModernBERT is the first open model to feature entire model unpadding and is the first encoder designed in a hardware-aware way to maximize inference efficiency.

ModernBERT pushes the encoder state of the art forward across a wide range of benchmarks. On GLUE, ModernBERT-base is the first encoder to beat DeBERTaV3-base since its release in 2021. ModernBERT is in a class of its own in code and ColBERT-style long-context retrieval benchmarks, scoring at least 6.85 and 9.1 percentage points higher than the closest model, respectively, while remaining state-of-the-art on short-context retrieval in both single and multi-vector settings. 

At the same time, ModernBERT processes short context inputs twice as fast as DeBERTaV3 and long-context inputs two times faster than the next fastest model with best-in-class memory efficiency.

ModernBERT is a generational leap over the original encoder models, with notable performance improvements over BERT and RoBERTa on both classification and retrieval tasks. ModernBERT is one of the few encoders to support long-context and programming applications, while simultaneously setting a new record in encoder inference efficiency.

\section{Limitations}

\textbf{Language} This study focuses exclusively on the English language, and trains on a very large number of tokens. As such, a major limitation of our work is that it is not directly applicable to other languages, and potentially even less-so to lower resources languages.

\textbf{Biases} Our model is trained largely on web data, as a result, all of its representations are subject to the biases present in such data.

\textbf{Harmful Content Generation} The MLM objective gives the model some ability to generate text by suggesting a given token to replace the [MASK] token~\cite{berticl}, which could result in the generation of harmful content. However, ModernBERT is not, primarily, a generative model, and as such, has not been trained to and therefore cannot generate longer sequences of text. As a result, it is considerably less likely to be at risk of generating harmful content of any kind.

\textbf{MLM-only objective}
Given the strong results of DeBERTav3 on classification tasks but weak ones on retrieval, it seems that a training leveraging both MLM and RTD might be better suited to achieve best results on classification. Extending our work to RTD is thus a promising line of research.

\textbf{Scaling}
Besides the architectural modifications, a key aspect of our studies is data scaling. However, other scaling axes, notably in terms of model parameters are left unexplored.

\section{Acknowledgements}
The authors would like to acknowledge \& thank the many people who assisted, supported, or offered insights useful for the completion of this project.

We are particularly thankful for the one-off implementation or evaluation work conducted by Jack Cook, Mark Tenenholtz, Johno Whitaker, and Wayde Gilliam. We also extend similar thanks to Zach Nussbaum for assisting in resolving issues we encountered with NomicBERT during evaluation.

We would like to acknowledge Enrico Shippole, Daniel Han, Colin Raffel, Pierre-Carl Langlais, Omar Khattab, Urchade Zaratiana, Aurélien Lac, Amélie Chatelain, and Raphaël Sourty, for their helpful contributions to discussions.

We also thank Weights\&Biases for providing free access to their platform, in particular Morgan McGuire and Thomas Capelle for their support.

We thank HuggingFace's Arthur Zucker, Cyril Vallez, and Pedro Cuenca for assisting with day-one HuggingFace support.

Finally, we acknowledge Orange Business Cloud Avenue as compute provider and their hardware support throughout the project and thank LightOn for sponsoring the compute.

\section{Contribution Statement}
BW, AC, and BC jointly led the project and contributed to all parts of it. \\
BW worked on all aspects of the project and contributed to all major decisions. He led model design, model training, implemented the majority of the model architecture, and assisted with data selection, elevations, and paper writing.\\
AC co-initiated the project and worked on all aspects of it, including project coordination. Notably, he contributed to monitoring training runs and co-led ablations, final evaluations and paper writing.\\
BC initiated the project and worked on all aspects of it. He contributed to model design and co-led final evaluations, led paper writing, and contributed to the context extension data processing.\\
OW led and conducted the majority of the data selection, processing, and discussion, for all stages of training. He also contributed valuable inputs throughout all stages of the project.\\
OH and ST contributed to a majority of the stages of the project, in particular model architecture and training, with both discussions, implementations and paper writing. Other contributions include pretraining monitoring, final traditional evaluations, and ablations. ST specifically worked on adapting the RoPE kernel for unpadded sequences and running the final GLUE benchmarks. OH additionally conducted a thorough investigation into complex issues that arose during training.\\
RB contributed greatly to the initial evaluation work, focusing on ablations and in-training evals.\\
AG and FL contributed to training efficiency, especially in implementing sequence packing.\\
AG and GA contributed to model evaluations, especially in long context evaluations.\\
TA contributed to discussions throughout the project and assisted in integrating the original research implementation with open source software.\\
NC contributed to context extension data mixtures, and provided insight into model training and on improving the quality of code data.\\
IP and JH provided guidance and support throughout the project, especially on key decisions.

\bibliography{custom}

\appendix

\section{Training Settings}
\label{app:trainingsettings}

Detailed training settings can be found in Table~\ref{tab:trainingsettings}.

During training we used MNLI as a live evaluation, along with validation loss and token accuracy metrics on a 500 million randomly sampled sequences from the source datasets. 

We use \href{https://github.com/composer/composer}{Composer}~\cite{composer} as our training framework and 
\href{https://github.com/search?q=optimi&type=repositories}{optimī}~\cite{optimi} for our optimizer implementations.

\newcolumntype{C}{>{\centering\arraybackslash}X} 

\begin{table*}[!t]
\small
\centering
\begin{tabularx}{\textwidth}{lCCCCCC} 
    \toprule
    & \multicolumn{2}{c}{Pretraining Phase} & \multicolumn{2}{c}{Context Extension: Phase One} & \multicolumn{2}{c}{Context Extension: Phase Two} \\
    \cmidrule{2-3} \cmidrule(lr){4-5} \cmidrule(lr){6-7}
    & Base & Large & Base & Large & Base & Large \\
    \midrule
    Training Tokens & \multicolumn{2}{c}{1.719 trillion} & \multicolumn{2}{c}{250 billion} & \multicolumn{2}{c}{50 billion} \\
    Max Sequence Length & \multicolumn{2}{c}{1,024} & \multicolumn{2}{c}{8,192} & \multicolumn{2}{c}{8,192} \\
    \midrule
    Batch Size & 4,608 & 4,928 & 72 & 77 & 72 & 78 \\
    \hspace{3mm}Warmup (tokens) & 50 billion & 10 billion & - & - & - & - \\
    Microbatch Size & 96 & 56 & 12 & 7 & 12 & 6 \\
    \midrule
    Learning Rate & 8e-4 & 5e-4, 5e-5 & 3e-4 & 5e-5 & 3e-4 & 5e-5 \\
    \hspace{3mm}Schedule & \multicolumn{2}{c}{Trapezoidal} & - & - & \multicolumn{2}{c}{1-sqrt} \\
    \hspace{3mm}Warmup (tokens) & 3 billion & 2 billion & - & - & - & - \\
    \hspace{3mm}Decay (tokens) & - & - & - & - & \multicolumn{2}{c}{50 billion} \\
    Weight Decay & 1e-5 & 1e-5, 1e-6 & 1e-5 & 1e-6 & 1e-5 & 1e-6 \\
    \midrule
    Total Time (hours) & 194.2 & 425.3 & 39.9 & 80.7 & 11.5 & 21.7 \\
    Training Time (hours) & 191.1 & 420.4 & 36.3 & 75.1 & 7.5 & 15.3 \\
    \midrule
    Model Initialization & Megatron & From Base & - & - & - & - \\
    \midrule
    Dropout (attn out) & \multicolumn{6}{l}{\hspace{1.2mm}0.1} \\
    Dropout (all other layers) & \multicolumn{6}{l}{\hspace{1.2mm}0.0} \\
    \midrule
    Optimizer & \multicolumn{6}{l}{\hspace{1.2mm}StableAdamW} \\
    \hspace{1.2mm}Betas & \multicolumn{6}{l}{\hspace{1.2mm}(0.90, 0.98)} \\
    \hspace{1.2mm}Epsilon & \multicolumn{6}{l}{\hspace{1.2mm}1e-06} \\
    \midrule
    Training Hardware & \multicolumn{6}{l}{\hspace{1.2mm}8x H100} \\
    Training Strategy & \multicolumn{6}{l}{\hspace{1.2mm}Distributed DataParallel} \\
    Software Libraries & \multicolumn{6}{l}{\hspace{1.2mm}PyTorch 2.4.0, Cuda 12.4.0, Composer 0.24.1, Flash Attention 2.6.3, FA3 commit 32792d3} \\
    \bottomrule
\end{tabularx}
\caption{ModernBERT training settings. Dropout and below are shared across all phases.}  
\label{tab:trainingsettings}
\end{table*}

\subsection{Batch Size Schedule}
\label{subsec:batchsizewarmup}
Batch size warmup is a common-knowledge trick to speed up model training when working with medium to large batch sizes. Instead of "wasting" a full batch on updating the suboptimal initial weight distribution, we update the model weights on a gradually increasing batch size. Batch size warmup is usually longer than learning rate warmup, and can be thought of as providing a higher initial learning rate with a mini-learning rate decay to the defined learning rate schedule. We warmup ModernBERT's batch size from 768 to 4,608 over 50 billion tokens and from 448 to 4,928 over 10 billion tokens, for -base and -large, respectively, with an uneven token schedule so each batch size has the same number of update steps.

\subsection{Weight Tiling}
\label{subsec:weightiling}
Following the Phi family of models~\cite{phi15,phi2}, we initialized ModernBERT-large directly from ModernBERT-base's pretraining weights using center tiling and Gopher layer scaling~\cite{gopher}. Since Base's weight matrices are smaller than Large's, we centered Base' weights, accounting for each token embedding and attention head, then filled rest the of the weights using wraparound. Like Phi, we tested center initialization with random edge values and tiling from an edge, but both of these underperformed center tiling with wraparound. This weight initialization strategy greatly accelerates ModernBERT-large's initial training.

\subsection{Weight Decay}
\label{subsec:weightdecay}
We did not apply weight decay to the bias terms or normalization layers. Instead of PyTorch-style decoupled weight decay, we applied fully decoupled weight decay following \citet{adamw}.

\subsection{Final Checkpoints}
\label{subsec:finalcheckpoints}
Inspired by recent work showing that checkpoint averaging yields stronger final models~\cite{llama3,jacolbertv25}, we selected our final checkpoints by experimenting with various averaging methods and evaluating them on a subset of evaluation tasks. In no cases did Exponential Moving Average during annealing, as used by ~\citet{llama3}, result in stronger performance. ModernBERT-base is the result of averaging the 3 best performing annealing checkpoints with the final one. Averaging did not yield successful results on the large size, ModernBERT-Large model is the best performing annealing checkpoint.

\section{Model Design}
\label{app:modeldesign}

From ~\citet{codesigning}, in addition to setting attention heads as multiples of 64 and setting the embedding matrix as a power of 2 or multiple of 64, there are three model design choices to maximize performance (assuming float16 or bfloat16 computation):

\begin{table*}[!t]
\small
\centering
\begin{tabular}{lcc}
\toprule
 & Base & Large \\
\midrule
Vocabulary & 50,368 & 50,368 \\
Unused Tokens & 83 & 83 \\
Layers & 22 & 28 \\
Hidden Size & 768 & 1024 \\
Transformer Block & Pre-Norm & Pre-Norm \\
Activation Function & GeLU & GeLU\\
Linear Bias & False & False\\
Attention & Multi-head & Multi-head \\
Attention Heads & 12 & 16 \\
Global Attention & Every three layers & Every three layers \\
Local Attention Window & 128 & 128 \\
Intermediate Size & 1,152 & 2,624 \\
GLU Expansion & 2,304 & 5,248 \\
Normalization & LayerNorm & LayerNorm \\
Norm Epsilon & 1e-5 & 1e-5 \\
Norm Bias & False & False \\
RoPE theta & 160,000 & 160,000 \\
Local Attn RoPE theta & 10,000 & 10,000 \\
\bottomrule
\end{tabular}
\caption{ModernBERT model design}  
\label{tab:modeldesign}
\end{table*}

\begin{itemize}
\item \textbf{Tensor Core Requirement}: Weight matrix dimensions should be divisible by 64
\item \textbf{Tile Quantization}: Weight matrix is divisible into 128 × 256 blocks.
\item \textbf{Wave Quantization}: Number of blocks is divisible by the number of streaming multiprocessors (SM).
\end{itemize}

Given that we wanted to target good performance across multiple GPUs with a wide variety of SM counts, wave quantization is an impossible ask. So we selected a basket of GPUs (NVIDIA T4, A10, L4, RTX 3090, RTX 4090, A100, and H100) and calculated the approximate SM utilization for each by dividing the modulus blocks by the number of SMs. This appeared to be a decent performance heuristic in our spot checking. We then designed our models to maximize performance on the basket of GPUs, putting more weight on inference GPUs.

\section{Training Log}
\label{sec:traininglog}

\subsection{Sampling Issue} 
\label{subsec:samplingissue}
Our first pretraining run of ModernBERT-base ended in disaster as the loss exhibited a slow seesaw pattern before slowly diverging. Despite using PyTorch's distributed random sampler, training metrics suggested that the model was training on the dataset in a non-random order. Like the Olmo authors\footnote{We found a comment and GitHub issue about this in the Olmo codebase after resolving the issue ourselves.}, we determined that the PyTorch random sampler returns sequentially biased samples when the number of samples is somewhere between 500 million and 1 billion samples\footnote{We did not conduct a rigorous statistical analysis to determine exactly when this happens.}. We resolved this issue by replacing the PyTorch sampler with NumPy's PCG64DXSM random sampler.

\subsection{Large Rollback}
\label{subsec:largerollback}
We rolled back and restarted ModernBERT-large training at a lower learning rate of 5e-5 and lower weight decay of 1e-6 for the last 800 billion tokens. Prior to restarting training, large’s training loss, validation metrics, and live evaluations on MNLI had plateaued for a few hundred billion tokens at the higher 5e-4 learning rate. In contrast, ModernBERT-base showed a continuous, but diminishing, improvement on training loss, validation metrics, and live evaluations through the entire 1.719 trillion token training phase. This highlights one of the risks of training with a constant learning rate, other learning rate schedules can mitigate selecting a too high learning rate (or too small batch size) by lowering the learning rate throughout training.

\section{Architecture ablations}
\label{sec:archablations}
To select the updates to add in the ModernBERT architecture, we performed different ablations, except where stated, most ablations where ran at the 8-20 billion token scale:
\begin{itemize}
    \item We compared two GLU layers, GeGLU and SwiGLU. We find close to no difference between the two and choose to use GeGLU layers.
    \item Using different percentage of the head dimension for the RoPE dimension (50, 75, 100). Lower percentages gave slightly better results. However, the observed difference was minimal. As the ablations were conducted at a considerably smaller scale than the final training, we choose to err on the side of caution and opt to keep the dimension at 100 \%  to avoid potentially hindering the capabilities of the fully trained models.
    \item Both LayerNorm and RMSNorm yielded very similar results. While RMSNorm is theoretically faster, at the time this work was conducted, PyTorch did not have a native RMSNorm implementation, leading to eager-mode RMSNorm being the default implementation used for many users. To ensure ModernBERT has the highest possible out-of-the-box efficiency, we choose to use LayerNorm in the final models.
    \item We investigated using parallel attention to compute the MLP and attention matrices at the same time, which has been shown to increase processing speeds for larger model sizes~\cite{palm}. However, for models within our targe sizes and pre-training sequence length, the speed-up we observed was minimal while we encountered significant degradation in downstream performance. As such, we do not use parallel attention. It is however possible that larger encoders and/or larger sequence lengths might see a different trade-off.
    \item We explored the use of alternating global/local attention, with global attention every 3 layers and local attention over a 128 token sliding window otherwise. This setup yielded identical downstream performance when compared to the use of global attention in every layer, even at 100 billion tokens, while resulting in major speedups.
    \item We experimented with multiple tokenizers, before selecting our final one, based on a modified OLMo~\cite{olmo} tokenizer, which performed the best out of the recent tokenizers evaluated. Tokenizers from the BERT and RoBERTa generation of encoder models had competitive downstream performance on MNLI, but we theorized that their lack of recent training data and lack of code support would hinder downstream applications. Interestingly, we observed significant downstream performance degradation when using the Llama 2~\cite{llama2} tokenizer.

\end{itemize}
\section{Extended results}
\label{app:fullresults}
\subsection{Full GLUE results}
\label{subsec:fullglueresults}
The results for all the models each GLUE subsets are presented in Table~\ref{tab:fullglue}. The values for prior models are extracted from the literature. As mentioned in Section~\ref{subsubsec:evalglue}, we follow standard practice~\cite{roberta,mosaic,debertav3} and conduct an hyperparameter search on each subset. More specifically, we perform a sweep over learning rates in $[1e{-5}, 3e{-5}, 5e{-5}, 8e{-5}]$, weight decay in $[1e{-6}, 5e{-6}, 8e{-6}, 1e{-5}]$, and number of epochs in $[1, 2, 3]$ for tasks in SST-2, MNLI, and RTE, and $[2, 5, 10]$ for tasks in QNLI, QQP, CoLA, MRPC, and STS-B. The final values are detailed in Table ~\ref{tab:glue-hyperparams}. Early stopping is used for all the fine-tuning runs which reduces the overall fine-tuning time considerably. RTE MRPC and STS-B checkpoints are trained starting from the MNLI checkpoint.

\begin{table*}[!t]
\small
\centering
\begin{tabular}{llrrcccccccc}
\toprule
& & & & \multicolumn{2}{c}{Single Sentence} & \multicolumn{3}{c}{Paraphrase and Similarity} & \multicolumn{3}{c}{Natural Language Inference} \\
\cmidrule(lr){5-6} \cmidrule(lr){7-9} \cmidrule(lr){10-12}
& Model & Params & Seq. & CoLA & SST-2 & MRPC & STS-B & QQP & MNLI & QNLI & RTE \\
\midrule
\parbox[t]{3mm}{\multirow{7}{*}{\rotatebox[origin=c]{90}{Base}}}
& BERT${}^\beta$ & 110M & 512 & 59.0 & 93.1 & 89.5 & 89.4 & 91.4 & 85.4 & 91.6 & 78.2 \\
& RoBERTa${}^\alpha$ & 125M & 512 & 63.6 & 94.8 & 90.2 & 91.2 & 91.9 & 87.6 & 92.8 & 78.7 \\
& DeBERTav3${}^\epsilon$ & 183M & 512 & \textbf{69.2} & 95.6 & 89.5 & 91.6 & \textbf{92.4} & \textbf{90.0} & \textbf{94.0} & 83.8 \\
& MosaicBERT-128${}^\beta$ & 137M & 128 & 58.2 & 93.5 & 89.0 & 90.3 & 92.0 & 85.6 & 91.4 & 83.0 \\
& NomicBERT-2048${}^\gamma$ & 137M & 2048 & 50.0 & 93.0 & 88.0 & 90.0 & 92.0 & 86.0 & 92.0 & 82.0 \\
& GTE-en-MLM${}^\delta$ & 137M & 8192 & 57.0 & 93.4 & 92.1 & 90.2 & 88.8 & 86.7 & 91.9 & 84.8 \\
& ModernBERT & 149M & 8192 & 65.1 & \textbf{96.0} & \textbf{92.2} & \textbf{91.8} & 92.1 & 89.1 & 93.9 & \textbf{87.4} \\
\midrule
\parbox[t]{3mm}{\multirow{5}{*}{\rotatebox[origin=c]{90}{Large}}}
& BERT${}^\beta$ & 330M & 512 & 56.2 & 93.3 & 87.8 & 90.6 & 90.9 & 86.3 & 92.8 & 83.8 \\
& RoBERTa${}^\alpha$ & 355M & 512 & 68.0 & 96.4 & 90.9 & 92.4 & 92.2 & 90.2 & 94.7 & 86.6 \\
& DeBERTav3${}^\zeta$ & 434M & 512 & \textbf{75.3} & 96.9 & 92.2 & \textbf{93.0} & \textbf{93.3} & \textbf{91.8} & \textbf{96.0} & \textbf{92.7} \\
& GTE-en-MLM${}^\delta$ & 434M & 8192 & 60.4 & 95.1 & \textbf{93.5} & 91.4 & 89.2 & 89.2 & 93.9 & 88.1 \\
& ModernBERT & 395M & 8192 & 71.4 & \textbf{97.1} & 91.7 & 92.8 & 92.7 & 90.8 & 95.2 & 92.1 \\
\bottomrule
\end{tabular}
\caption{GLUE~\cite{wang-etal-2018-glue} dev set scores. $^\alpha$ taken from Table 8 of~\cite{roberta}, $^\beta$ taken from Table S3 of~\cite{mosaic}, $^\gamma$ from Table 2 of~\cite{nomic}, $^\delta$ from Table 21 of~\cite{gte}, $^\epsilon$ from Table 2 of~\cite{debertabaseresults} and $^\zeta$ from Table 3 of~\cite{debertav3}}
\label{tab:fullglue}
\end{table*}

\begin{table*}[]
\centering
\begin{tabular}{lcccccc}
\toprule
 & \multicolumn{3}{c}{\textbf{Base}} & \multicolumn{3}{c}{\textbf{Large}} \\
 \cmidrule(lr){2-4}  \cmidrule(lr){5-7} 
\textbf{Task} & LR & WD & Ep & LR & WD & Ep \\
\midrule
CoLA & $8e{-5}$ & $1e{-6}$ & 5 & $3e{-5}$ & $8e{-6}$ & 5 \\
MNLI & $5e{-5}$ & $5e{-6}$ & 1 & $3e{-5}$ & $1e{-5}$ & 1 \\
MRPC & $5e{-5}$ & $5e{-6}$ & 10 & $8e{-5}$ & $5e{-6}$ & 2 \\
QNLI & $8e{-5}$ & $5e{-6}$ & 2 & $3e{-5}$ & $5e{-6}$ & 2 \\
QQP & $5e{-5}$ & $5e{-6}$ & 10 & $5e{-5}$ & $8e{-6}$ & 2 \\
RTE & $5e{-5}$ & $1e{-5}$ & 3 & $5e{-5}$ & $8e{-6}$ & 3 \\
SST-2 & $8e{-5}$ & $1e{-5}$ & 2 & $1e{-5}$ & $1e{-6}$ & 3 \\
STSB & $8e{-5}$ & $5e{-6}$ & 10 & $8e{-5}$ & $1e{-5}$ & 10 \\
\bottomrule
\end{tabular}
\caption{Fine-tuning hyperparameters for ModernBERT on GLUE tasks. LR: Learning Rate, WD: Weight Decay, Ep: Epochs.}
\label{tab:glue-hyperparams}
\end{table*}

\subsection{Full BEIR results}
\label{subsec:fullBEIRresults}
In the main body, we only report the average score over the 15 very diverse datasets of BEIR. We report the results on every subsets for both single and multi-vector retrieval in Table~\ref{tab:fullbeirdpr} and Table~\ref{tab:fullbeircolbert} respectively.
For both settings and for every model, we perform a sweep for learning rates in $[1e{-5}, 2e{-5}, 3e{-5}, 5e{-5}, 8e{-5}, 1e{-4}]$ and choose the model obtaining the best average result over a subset of datasets composed of NFCorpus, SciFact, TREC-Covid and FiQA as the final model. Best learning rates for every setting are reported in Table~\ref{tab:retrieval-hyperparams}. Although ModernBERT showcase strong results across the board, it should be noted that an important factor in its performance is TREC-COVID~\cite{treccovid}, potentially showcasing the benefits of ModernBERT being trained with a more recent knowledge cutoff than most existing encoders. However, NomicBERT and GTE have also been trained on updated data, so the cutoff cannot be the only factor affecting the performance.

\begin{table*}[!t]
\resizebox{\textwidth}{!}{
\centering
\begin{tabular}{ll|ccccccccccccccc|r}
\toprule
& Model & NFCorpus & SciFact & TREC-Covid & FiQA & ArguAna & Climate-FEVER & DBPedia & FEVER & HotpotQA & MSMARCO & NQ & Quora & SciDocs & Touche2020 & CQADupstack & Avg. \\
\midrule
\parbox[t]{3mm}{\multirow{6}{*}{\rotatebox[origin=c]{90}{Base}}}
& BERT & 24.3 & 51.3 & 49.5 & 22.8 & 31.6 & 21.9 & 28.2 & 64.1 & 47.9 & 58.5 & 37.9 & 83.1 & 12.9 & 20.4 & 28.5 & 38.9 \\
& RoBERTa & 20.4 & 45.6 & 52.2 & 26.1 & 35.2 & 22.3 & 23.1 & 60.2 & 45.0 & 56.0 & 34.7 & 84.0 & 11.4 & \textbf{21.1} & 28.8 & 37.7 \\
& DeBERTaV3 & 8.0 & 22.6 & 48.4 & 11.5 & 26.1 & 9.7 & 5.3 & 17.3 & 8.0 & 25.2 & 12.5 & 74.7 & 5.4 & 14.2 & 14.2 & 20.2 \\
& NomicBERT & 25.7 & 52.0 & 63.0 & 23.5 & 35.5 & 22.9 & \textbf{30.3} & 65.0 & 48.0 & 60.6 & \textbf{42.6} & 84.5 & 12.6 & 19.0 & 29.2 & 41.0 \\
& GTE-en-MLM & \textbf{26.3} & 54.1 & 49.7 & \textbf{30.1} & \textbf{35.7} & \textbf{24.5} & 28.9 & \textbf{66.5} & \textbf{49.9} & \textbf{63.1} & 41.7 & 85.2 & \textbf{14.1} & 19.1 & 32.5 & 41.4 \\
& ModernBERT & 23.7 & \textbf{57.0} & \textbf{72.1} & 28.8 & 35.7 & 23.6 & 23.8 & 59.9 & 46.1 & 61.6 & 39.5 & \textbf{85.9} & 12.5 & 20.8 & \textbf{33.1} & \textbf{41.6} \\
\midrule
\parbox[t]{3mm}{\multirow{5}{*}{\rotatebox[origin=c]{90}{Large}}}
& BERT & 23.3 & 50.7 & 48.9 & 24.0 & 35.2 & 22.1 & \textbf{27.2} & 61.7 & 45.9 & 59.8 & 39.5 & 83.6 & 13.0 & 19.5 & 28.9 & 38.9 \\
& RoBERTa & 23.9 & 53.4 & 55.0 & 33.4 & 37.6 & \textbf{23.5} & 25.4 & 65.2 & 47.1 & 60.4 & 43.3 & 85.8 & 13.7 & 21.1 & 33.0 & 41.4 \\
& DeBERTaV3 & 9.6 & 31.2 & 56.6 & 15.8 & 26.3 & 14.4 & 6.8 & 29.4 & 15.3 & 32.4 & 21.5 & 79.1 & 7.0 & 18.8 & 19.9 & 25.6 \\
& GTE-en-MLM & \textbf{27.7} & 57.6 & 48.4 & \textbf{34.0} & 35.3 & 24.0 & 27.0 & \textbf{65.4} & \textbf{50.8} & 64.1 & 44.9 & 85.3 & \textbf{15.6} & 21.4 & 35.5 & 42.5 \\
& ModernBERT & 26.2 & \textbf{60.4} & \textbf{74.1} & 33.1 & \textbf{38.2} & 20.5 & 25.1 & 62.7 & 49.2 & \textbf{64.9} & \textbf{45.5} & \textbf{86.5} & 13.8 & \textbf{23.1} & \textbf{36.5} & \textbf{44.0} \\
\bottomrule
\end{tabular}
}
\caption{BEIR~\cite{BEIR} nDCG@10 scores for single-vector retrieval models.}
\label{tab:fullbeirdpr}
\end{table*}

\begin{table*}[!t]
\resizebox{\textwidth}{!}{
\centering
\begin{tabular}{ll|ccccccccccccccc|r}
\toprule
& Model & NFCorpus & SciFact & TREC-Covid & FiQA & ArguAna & Climate-FEVER & DBPedia & FEVER & HotpotQA & MSMARCO & NQ & Quora & SciDocs & Touche2020 & CQADupstack & Avg. \\
\midrule
\parbox[t]{3mm}{\multirow{6}{*}{\rotatebox[origin=c]{90}{Base}}}
& BERT & 34.2 & 71.5 & 69.9 & 35.0 & \textbf{49.9} & 19.2 & 42.4 & 83.1 & 69.8 & 45.4 & 55.4 & 84.1 & 14.7 & 27.0 & 34.2 & 49.0 \\
& RoBERTa & 33.7 & 70.8 & 69.8 & 37.4 & 48.9 & 18.9 & 39.3 & 81.2 & 66.1 & 43.7 & 56.3 & 83.6 & 14.8 & 31.7 & 34.4 & 48.7 \\
& DeBERTaV3 & 31.9 & 68.5 & 75.5 & 35.5 & 46.5 & 18.3 & 35.6 & 78.1 & 65.3 & 39.5 & 50.4 & 83.7 & 14.6 & 31.1 & 32.3 & 47.1 \\
& NomicBERT & \textbf{35.5} & 72.2 & 73.5 & 35.9 & 44.8 & 19.0 & \textbf{43.6} & 83.9 & \textbf{71.1} & \textbf{46.3} & \textbf{58.5} & 84.0 & 15.1 & 31.3 & 33.9 & 49.9 \\
& GTE-en-MLM & 35.1 & 71.5 & 69.4 & 36.0 & 48.5 & 17.4 & 41.2 & 79.9 & 67.0 & 44.4 & 52.8 & 85.2 & 15.0 & 25.4 & 34.6 & 48.2 \\
& ModernBERT & 35.2 & \textbf{73.0} & \textbf{80.5} & \textbf{38.0} & 49.1 & \textbf{22.2} & 42.0 & \textbf{85.8} & 70.4 & 45.4 & 57.1 & \textbf{86.3} & \textbf{16.0} & \textbf{33.9} & \textbf{35.1} & \textbf{51.3} \\
\midrule
\parbox[t]{3mm}{\multirow{5}{*}{\rotatebox[origin=c]{90}{Large}}}
& BERT & 34.6 & 72.9 & 68.8 & 35.5 & 48.3 & 19.7 & 42.4 & 83.6 & 70.7 & 45.9 & 57.2 & 84.8 & 15.2 & 28.9 & 34.9 & 49.5 \\
& RoBERTa & 35.0 & 72.3 & 74.4 & 38.7 & 50.0 & 19.6 & 41.0 & 82.0 & 66.2 & 44.7 & 57.5 & 85.9 & 15.3 & 27.9 & \textbf{36.0} & 49.8 \\
& DeBERTaV3 & 31.7 & 70.2 & 73.3 & 35.0 & 46.2 & 18.0 & 36.5 & 79.0 & 63.2 & 39.4 & 51.6 & 81.1 & 14.1 & 28.6 & 33.1 & 46.7 \\
& GTE-en-MLM & 35.2 & 72.4 & 67.2 & 39.6 & 50.3 & 20.8 & \textbf{44.4} & 82.5 & 72.0 & \textbf{47.0} & \textbf{60.1} & \textbf{86.4} & 15.9 & 30.9 & 35.4 & 50.7 \\
& ModernBERT & \textbf{36.0} & \textbf{73.2} & \textbf{81.3} & \textbf{40.3} & \textbf{50.3} & \textbf{22.3} & 44.1 & \textbf{85.8} & \textbf{72.5} & 46.0 & 59.9 & 86.1 & \textbf{16.9} & \textbf{34.6} & 35.9 & \textbf{52.4} \\
\bottomrule
\end{tabular}
}
\caption{BEIR~\cite{BEIR} nDCG@10 scores for multi-vector retrieval models.}
\label{tab:fullbeircolbert}
\end{table*}

\begin{table*}[!t]
\centering
\begin{tabular}{llcc}
\toprule
& Model & Single-vector (DPR) & Multi-vector (ColBERT) \\
\midrule
\parbox[t]{3mm}{\multirow{6}{*}{\rotatebox[origin=c]{90}{Base}}}
& BERT & $5\times10^{-5}$ & $8\times10^{-5}$ \\
& RoBERTa & $3\times10^{-5}$ & $8\times10^{-5}$ \\
& DeBERTaV3 & $8\times10^{-5}$ & $5\times10^{-5}$ \\
& NomicBERT & $5\times10^{-5}$ & $1\times10^{-4}$ \\
& GTE-en-MLM & $5\times10^{-5}$ & $8\times10^{-5}$ \\
& ModernBERT & $8\times10^{-5}$ & $1\times10^{-4}$ \\
\midrule
\parbox[t]{3mm}{\multirow{5}{*}{\rotatebox[origin=c]{90}{Large}}}
& BERT & $3\times10^{-5}$ & $1\times10^{-4}$ \\
& RoBERTa & $3\times10^{-5}$ & $1\times10^{-5}$ \\
& DeBERTaV3 & $8\times10^{-5}$ & $1\times10^{-5}$ \\
& GTE-en-MLM & $3\times10^{-5}$ & $3\times10^{-5}$ \\
& ModernBERT & $1\times10^{-4}$ & $3\times10^{-5}$ \\
\bottomrule
\end{tabular}
\caption{Learning rate used for reported results on BEIR~\cite{BEIR} for both single and multi vector retrieval}
\label{tab:retrieval-hyperparams}
\end{table*}

\section{Efficiency}
\label{app:efficiency}
Full statistics of the synthetic datasets used to evaluate the efficiency of the models in Section~\ref{sec:efficiencyresults} are given in Table~\ref{tab:syntheticdatasets}. The detailed runtimes, alongside with the maximum batch size for every model is detailed in Table~\ref{tab:efficiencyfulltable}.

The high maximum batch-size achieved by ModernBERT models, considerably higher than any other models, highlight the strong memory efficiency of the model at both sizes. Inversely, it is worth noting that while DeBERTaV3 has competitive GLUE performance, it stands out as particularly inefficient, both in its memory use and processing speed. Indeed, on both model sizes, DeBERTaV3's memory use is 5-to-7 times higher than ModernBERT's, and it processes inputs, two times slower even in the most favorable scenario where all sequences are at the maximum possible length, thus negating any advantage from unpadding.

\begin{table*}[!t]
\centering
\begin{tabular}{lrrrr}
\toprule
 & \multicolumn{2}{c}{Short} & \multicolumn{2}{c}{Long} \\
\cmidrule(lr){2-3} \cmidrule(lr){4-5}
& Fixed & Variable & Fixed & Variable \\
\midrule
Total Token Count & 4,194,304 & 2,096,510 & 67,108,864 & 33,604,913 \\
Standard deviation & 0 & 64 & 0 & 1,024 \\
Average Length & 512 & 256 & 8,192 & 4,102 \\
Longest sequence & 512 & 476 & 8,192 & 7,624 \\
Shortest sequence & 512 & 32 & 8,192 & 171 \\
Number of sequences & 8,192 & 8,192 & 8,192 & 8,192 \\
\bottomrule
\end{tabular}
\caption{Token statistics for the synthetic datasets used in efficiency evaluations.}
\label{tab:syntheticdatasets}
\end{table*}

\begin{table*}[!t]
\resizebox{\textwidth}{!}{
\centering
\begin{tabular}{llrrrrrrr}
\toprule
& & & \multicolumn{3}{c}{Short} & \multicolumn{3}{c}{Long} \\
\cmidrule(lr){4-6} \cmidrule(lr){7-9}
& Model & Params & BS & Fixed & Variable & BS & Fixed & Variable \\
\midrule
\parbox[t]{3mm}{\multirow{7}{*}{\rotatebox[origin=c]{90}{Base}}}
& BERT & 110M & 1096 & \textbf{23.3 ± 0.02} & -- & -- & -- & -- \\
& RoBERTa & 125M & 664 & \textbf{23.3 ± 0.19} & -- & -- & -- & -- \\
& DeBERTaV3 & 183M & 236 & 59.7 ± 0.11 & -- & -- & -- & -- \\
& NomicBERT & 137M & 588 & 35.8 ± 0.01 & -- & 36 & 1455.5 ± 0.31 & -- \\
& GTE-en-MLM & 137M & 640 & 33.9 ± 1.21 & -- & 38 & 1434.7 ± 3.69 & -- \\
& GTE-en-MLM\textsubscript{xformers} & 137M & 640 & 34.2 ± 0.10 & 16.3 ± 0.04 & 38 & 1412.6 ± 3.19 & 499.2 ± 0.11 \\
& ModernBERT & 149M & \textbf{1604} & 28.3 ± 0.55 & \textbf{14.2 ± 0.01} & \textbf{98} & \textbf{542.4 ± 0.20} & \textbf{251.2 ± 0.32} \\
\midrule
\parbox[t]{3mm}{\multirow{6}{*}{\rotatebox[origin=c]{90}{Large}}}
& BERT & 330M & \textbf{792} & \textbf{77.1 ± 1.50} & -- & -- & -- & -- \\
& RoBERTa & 355M & 460 & 99.8 ± 1.79 & -- & -- & -- & -- \\
& DeBERTaV3 & 434M & 134 & 170.8 ± 0.06 & -- & -- & -- & -- \\
& GTE-en-MLM & 435M & 472 & 108.4 ± 0.07 & -- & 28 & 4144.7 ± 0.05 & -- \\
& GTE-en-MLM\textsubscript{xformers} & 435M & 472 & 109.0 ± 0.14 & 51.9 ± 0.02 & 28 & 4059.1 ± 4.55 & 1476.3 ± 0.94 \\
& ModernBERT & 395M & 770 & \textbf{80.1 ± 1.65} & \textbf{39.6 ± 0.02} & \textbf{48} & \textbf{1433.9 ± 0.99} & \textbf{674.9 ± 0.15} \\
\bottomrule
\end{tabular}
}
\caption{Inference runtime for all models. Bold indicates the best for the column within two SDs.}
\label{tab:efficiencyfulltable}
\end{table*}

\section{Licensing}

We release the ModernBERT model architecture, model weights, and training codebase under the Apache 2.0 license.

\end{document}